# Integrating Multiple Knowledge Sources to Disambiguate Word Sense: An Exemplar-Based Approach

**Hwee Tou Ng**
Defence Science Organisation
20 Science Park Drive
Singapore 118230
nhweetou@trantor.dso.gov.sg

**Hian Beng Lee**
Defence Science Organisation
20 Science Park Drive
Singapore 118230
lhianben@trantor.dso.gov.sg

## Abstract

In this paper, we present a new approach for word sense disambiguation (WSD) using an exemplar-based learning algorithm. This approach integrates a diverse set of knowledge sources to disambiguate word sense, including part of speech of neighboring words, morphological form, the unordered set of surrounding words, local collocations, and verb-object syntactic relation. We tested our WSD program, named LEXAS, on both a common data set used in previous work, as well as on a large sense-tagged corpus that we separately constructed. LEXAS achieves a higher accuracy on the common data set, and performs better than the most frequent heuristic on the highly ambiguous words in the large corpus tagged with the refined senses of WORD-NET.

## 1 Introduction

One important problem of Natural Language Processing (NLP) is figuring out what a word means when it is used in a particular context. The different meanings of a word are listed as its various senses in a dictionary. The task of Word Sense Disambiguation (WSD) is to identify the correct sense of a word in context. Improvement in the accuracy of identifying the correct word sense will result in better machine translation systems, information retrieval systems, etc. For example, in machine translation, knowing the correct word sense helps to select the appropriate target words to use in order to translate into a target language.

In this paper, we present a new approach for WSD using an exemplar-based learning algorithm. This approach integrates a diverse set of knowledge sources to disambiguate word sense, including part of speech (POS) of neighboring words, morphological form, the unordered set of surrounding words, local collocations, and verb-object syntactic relation. To evaluate our WSD program, named LEXAS (LEXical Ambiguity-resolving System), we tested it on a common data set involving the noun "interest" used by Bruce and Wiebe (Bruce and Wiebe, 1994). LEXAS achieves a mean accuracy of 87.4% on this data set, which is higher than the accuracy of 78% reported in (Bruce and Wiebe, 1994).

Moreover, to test the scalability of LEXAS, we have acquired a corpus in which 192,800 word occurrences have been manually tagged with senses from WORD-NET, which is a public domain lexical database containing about 95,000 word forms and 70,000 lexical concepts (Miller, 1990). These sense tagged word occurrences consist of 191 most frequently occurring and most ambiguous nouns and verbs. When tested on this large data set, LEXAS performs better than the default strategy of picking the most frequent sense. To our knowledge, this is the first time that a WSD program has been tested on such a large scale, and yielding results better than the most frequent heuristic on highly ambiguous words with the refined sense distinctions of WORDNET.

## 2 Task Description

The input to a WSD program consists of unrestricted, real-world English sentences. In the output, each word occurrence $w$ is tagged with its correct sense (according to the context) in the form of a sense number $i$, where $i$ corresponds to the $i$-th sense definition of $w$ as given in some dictionary. The choice of which sense definitions to use (and according to which dictionary) is agreed upon in advance.

For our work, we use the sense definitions as given in WORDNET, which is comparable to a good desktop printed dictionary in its coverage and sense distinction. Since WORDNET only provides sense definitions for content words, (i.e., words in the parts of speech (POS) noun, verb, adjective, and adverb), LEXAS is only concerned with disambiguating the sense of content words. However, almost all existing work in WSD deals only with disambiguating content words too.

LEXAS assumes that each word in an input sentence has been pre-tagged with its correct POS, so that the possible senses to consider for a content word $w$ are only those associated with the particular POS of $w$ in the sentence. For instance, given the sentence "A reduction of principal and interest is one way the problem may be solved.", since the word "interest" appears as a noun in this sentence, LEXAS will only consider the noun senses of "interest" but not its verb senses. That is, LEXAS is only concerned with disambiguating senses of a word in a given POS. Making such an assumption is reasonable since POS taggers that can achieve accuracy of 96% are readily available to assign POS to unrestricted English sentences (Brill, 1992; Cutting et al., 1992).

In addition, sense definitions are only available for root words in a dictionary. These are words that are not morphologically inflected, such as "interest" (as opposed to the plural form "interests"), "fall" (as opposed to the other inflected forms like "fell", "fallen", "falling", "falls"), etc. The sense of a morphologically inflected content word is the sense of its uninflected form. LEXAS follows this convention by first converting each word in an input sentence into its morphological root using the morphological analyzer of WORDNET, before assigning the appropriate word sense to the root form.

## 3 Algorithm

LEXAS performs WSD by first learning from a training corpus of sentences in which words have been pre-tagged with their correct senses. That is, it uses supervised learning, in particular exemplar-based learning, to achieve WSD. Our approach has been fully implemented in the program LEXAS. Part of the implementation uses PEBLS (Cost and Salzberg, 1993; Rachlin and Salzberg, 1993), a public domain exemplar-based learning system.

LEXAS builds one exemplar-based classifier for each content word $w$. It operates in two phases: training phase and test phase. In the training phase, LEXAS is given a set $S$ of sentences in the training corpus in which sense-tagged occurrences of $w$ appear. For each training sentence with an occurrence of $w$, LEXAS extracts the parts of speech (POS) of words surrounding $w$, the morphological form of $w$, the words that frequently co-occur with $w$ in the same sentence, and the local collocations containing $w$. For disambiguating a noun $w$, the verb which takes the current noun $w$ as the object is also identified. This set of values form the features of an example, with one training sentence contributing one training example.

Subsequently, in the test phase, LEXAS is given new, previously unseen sentences. For a new sentence containing the word $w$, LEXAS extracts from the new sentence the values for the same set of features, including parts of speech of words surrounding $w$, the morphological form of $w$, the frequently co-occurring words surrounding $w$, the local collocations containing $w$, and the verb that takes $w$ as an object (for the case when $w$ is a noun). These values form the features of a test example.

This test example is then compared to every training example. The sense of word $w$ in the test example is the sense of $w$ in the closest matching training example, where there is a precise, computational definition of "closest match" as explained later.

### 3.1 Feature Extraction

The first step of the algorithm is to extract a set $F$ of features such that each sentence containing an occurrence of $w$ will form a training example supplying the necessary values for the set $F$ of features.

Specifically, LEXAS uses the following set of features to form a training example:

$$L_3, L_2, L_1, R_1, R_2, R_3, M, K_1, \ldots, K_m, C_1, \ldots, C_9, V$$

#### 3.1.1 Part of Speech and Morphological Form

The value of feature $L_i$ is the part of speech (POS) of the word i-th position to the left of $w$. The value of $R_i$ is the POS of the word i-th position to the right of $w$. Feature $M$ denotes the morphological form of $w$ in the sentence $s$. For a noun, the value for this feature is either singular or plural; for a verb, the value is one of infinitive (as in the uninflected form of a verb like "fall"), present-third-person-singular (as in "falls"), past (as in "fell"), present-participle (as in "falling") or past-participle (as in "fallen").

#### 3.1.2 Unordered Set of Surrounding Words

$K_1, \ldots, K_m$ are features corresponding to a set of keywords that frequently co-occur with word $w$ in the same sentence. For a sentence $s$, the value of feature $K_i$ is one if the keyword $K_i$ appears somewhere in sentence $s$, else the value of $K_i$ is zero.

The set of keywords $K_1, \ldots, K_m$ are determined based on conditional probability. All the word tokens other than the word occurrence $w$ in a sentence $s$ are candidates for consideration as keywords. These tokens are converted to lower case form before being considered as candidates for keywords.

Let $cp(i|k)$ denotes the conditional probability of sense $i$ of $w$ given keyword $k$, where

$$cp(i|k) = \frac{N_{i,k}}{N_k}$$

$N_k$ is the number of sentences in which keyword $k$ co-occurs with $w$, and $N_{i,k}$ is the number of sentences in which keyword $k$ co-occurs with $w$ where $w$ has sense $i$.

For a keyword $k$ to be selected as a feature, it must satisfy the following criteria:

1. $cp(i|k) \geq M_1$ for some sense $i$, where $M_1$ is some predefined minimum probability.

2. The keyword $k$ must occur at least $M_2$ times in some sense $i$, where $M_2$ is some predefined minimum value.

3. Select at most $M_3$ number of keywords for a given sense $i$ if the number of keywords satisfying the first two criteria for a given sense $i$ exceeds $M_3$. In this case, keywords that co-occur more frequently (in terms of absolute frequency) with sense $i$ of word $w$ are selected over those co-occurring less frequently.

Condition 1 ensures that a selected keyword is indicative of some sense $i$ of $w$ since $cp(i|k)$ is at least some minimum probability $M_1$. Condition 2 reduces the possibility of selecting a keyword based on spurious occurrence. Condition 3 prefers keywords that co-occur more frequently if there is a large number of eligible keywords.

For example, $M_1 = 0.8$, $M_2 = 5$, $M_3 = 5$ when LEXAS was tested on the common data set reported in Section 4.1.

To illustrate, when disambiguating the noun "interest", some of the selected keywords are: expressed, acquiring, great, attracted, expressions, pursue, best, conflict, served, short, minority, rates, rate, bonds, lower, payments.

### 3.1.3 Local Collocations

Local collocations are common expressions containing the word to be disambiguated. For our purpose, the term *collocation* does not imply idiomatic usage, just words that are frequently adjacent to the word to be disambiguated. Examples of local collocations of the noun "interest" include "in the interest of", "principal and interest", etc. When a word to be disambiguated occurs as part of a collocation, its sense can be frequently determined very reliably. For example, the collocation "in the interest of" always implies the "advantage, advancement, favor" sense of the noun "interest". Note that the method for extraction of keywords that we described earlier will fail to find the words "in", "the", "of" as keywords, since these words will appear in many different positions in a sentence for many senses of the noun "interest". It is only when these words appear in the exact order "in the interest of" around the noun "interest" that strongly implies the "advantage, advancement, favor" sense.

| Left Offset | Right Offset | Collocation Example |
|---|---|---|
| -1 | -1 | accrued interest |
| 1 | 1 | interest rate |
| -2 | -1 | principal and interest |
| -1 | 1 | national interest in |
| 1 | 2 | interest and dividends |
| -3 | -1 | sale of an interest |
| -2 | 1 | in the interest of |
| -1 | 2 | an interest in a |
| 1 | 3 | interest on the bonds |

Table 1: Features for Collocations

There are nine features related to collocations in an example. Table 1 lists the nine features and some collocation examples for the noun "interest". For example, the feature with left offset = -2 and right offset = 1 refers to the possible collocations beginning at the word two positions to the left of "interest" and ending at the word one position to the right of "interest". An example of such a collocation is "in the interest of".

The method for extraction of local collocations is similar to that for extraction of keywords. For each of the nine collocation features, LEXAS concatenates the words between the left and right offset positions. Using similar conditional probability criteria for the selection of keywords, collocations that are predictive of a certain sense are selected to form the possible values for a collocation feature.

### 3.1.4 Verb-Object Syntactic Relation

LEXAS also makes use of the verb-object syntactic relation as one feature $V$ for the disambiguation of nouns. If a noun to be disambiguated is the head of a noun group, as indicated by its last position in a noun group bracketing, and if the word immediately preceding the opening noun group bracketing is a verb, LEXAS takes such a verb-noun pair to be in a verb-object syntactic relation. Again, using similar conditional probability criteria for the selection of keywords, verbs that are predictive of a certain sense of the noun to be disambiguated are selected to form the possible values for this verb-object feature $V$.

Since our training and test sentences come with noun group bracketing, determining verb-object relation using the above heuristic can be readily done. In future work, we plan to incorporate more syntactic relations including subject-verb, and adjective-headnoun relations. We also plan to use verb-object and subject-verb relations to disambiguate verb senses.

### 3.2 Training and Testing

The heart of exemplar-based learning is a measure of the similarity, or distance, between two examples. If the distance between two examples is small, then the two examples are similar. We use the following definition of distance between two symbolic values $v_1$ and $v_2$ of a feature $f$:

$$d(v_1, v_2) = \sum_{i=1}^{n} |\frac{C_{1,i}}{C_1} - \frac{C_{2,i}}{C_2}|$$

$C_{1,i}$ is the number of training examples with value $v_1$ for feature $f$ that is classified as sense $i$ in the training corpus, and $C_1$ is the number of training examples with value $v_1$ for feature $f$ in any sense. $C_{2,i}$ and $C_2$ denote similar quantities for value $v_2$ of feature $f$. $n$ is the total number of senses for a word $w$.

This metric for measuring distance is adopted from (Cost and Salzberg, 1993), which in turn is adapted from the value difference metric of the earlier work of (Stanfill and Waltz, 1986). The distance between two examples is the sum of the distances between the values of all the features of the two examples.

During the training phase, the appropriate set of features is extracted based on the method described in Section 3.1. From the training examples formed, the distance between any two values for a feature $f$ is computed based on the above formula.

During the test phase, a test example is compared against *all* the training examples. LEXAS then determines the closest matching training example as the one with the minimum distance to the test example. The sense of $w$ in the test example is the sense of $w$ in this closest matching training example.

If there is a tie among several training examples with the same minimum distance to the test example, LEXAS randomly selects one of these training examples as the closet matching training example in order to break the tie.

## 4 Evaluation

To evaluate the performance of LEXAS, we conducted two tests, one on a common data set used in (Bruce and Wiebe, 1994), and another on a larger data set that we separately collected.

| LDOCE sense | Frequency | Percent |
|---|---|---|
| 1: readiness to give attention | 361 | 15% |
| 2: quality of causing attention to be given | 11 | <1% |
| 3: activity, subject, etc. which one gives time and attention to | 67 | 3% |
| 4: advantage, advancement, or favor | 178 | 8% |
| 5: a share (in a company, business, etc.) | 499 | 21% |
| 6: money paid for the use of money | 1253 | 53% |

Table 2: Distribution of Sense Tags

### 4.1 Evaluation on a Common Data Set

To our knowledge, very few of the existing work on WSD has been tested and compared on a common data set. This is in contrast to established practice in the machine learning community. This is partly because there are not many common data sets publicly available for testing WSD programs.

One exception is the sense-tagged data set used in (Bruce and Wiebe, 1994), which has been made available in the public domain by Bruce and Wiebe. This data set consists of 2369 sentences each containing an occurrence of the noun "interest" (or its plural form "interests") with its correct sense manually tagged. The noun "interest" occurs in six different senses in this data set. Table 2 shows the distribution of sense tags from the data set that we obtained. Note that the sense definitions used in this data set are those from Longman Dictionary of Contemporary English (LDOCE) (Procter, 1978). This does not pose any problem for LEXAS, since LEXAS only requires that there be a division of senses into different classes, regardless of how the sense classes are defined or numbered.

POS of words are given in the data set, as well as the bracketings of noun groups. These are used to determine the POS of neighboring words and the verb-object syntactic relation to form the features of examples.

In the results reported in (Bruce and Wiebe, 1994), they used a test set of 600 randomly selected sentences from the 2369 sentences. Unfortunately, in the data set made available in the public domain, there is no indication of which sentences are used as test sentences. As such, we conducted 100 random trials, and in each trial, 600 sentences were randomly selected to form the test set. LEXAS is trained on the remaining 1769 sentences, and then tested on a *separate* test set of sentences in each trial.

| WSD research | Accuracy |
|---|---|
| Black (1988) | 72% |
| Zernik (1990) | 70% |
| Yarowsky (1992) | 72% |
| Bruce & Wiebe (1994) | 79% |
| LEXAS (1996) | 89% |

Table 3: Comparison with previous results

| Knowledge Source | Mean Accuracy | Std Dev |
|---|---|---|
| POS & morpho | 77.2% | 1.44% |
| surrounding words | 62.0% | 1.82% |
| collocations | 80.2% | 1.55% |
| verb-object | 43.5% | 1.79% |

Table 4: Relative Contribution of Knowledge Sources

Note that in Bruce and Wiebe's test run, the proportion of sentences in each sense in the test set is approximately equal to their proportion in the whole data set. Since we use random selection of test sentences, the proportion of each sense in our test set is also approximately equal to their proportion in the whole data set in our random trials.

The average accuracy of LEXAS over 100 random trials is 87.4%, and the standard deviation is 1.37%. In each of our 100 random trials, the accuracy of LEXAS is always higher than the accuracy of 78% reported in (Bruce and Wiebe, 1994).

Bruce and Wiebe also performed a separate test by using a subset of the "interest" data set with only 4 senses (sense 1, 4, 5, and 6), so as to compare their results with previous work on WSD (Black, 1988; Zernik, 1990; Yarowsky, 1992), which were tested on 4 senses of the noun "interest". However, the work of (Black, 1988; Zernik, 1990; Yarowsky, 1992) were not based on the present set of sentences, so the comparison is only suggestive. We reproduced in Table 3 the results of past work as well as the classification accuracy of LEXAS, which is 89.9% with a standard deviation of 1.09% over 100 random trials.

In summary, when tested on the noun "interest", LEXAS gives higher classification accuracy than previous work on WSD.

In order to evaluate the relative contribution of the knowledge sources, including (1) POS and morphological form; (2) unordered set of surrounding words; (3) local collocations; and (4) verb to the left (verb-object syntactic relation), we conducted 4 separate runs of 100 random trials each. In each run, we utilized only one knowledge source and compute the average classification accuracy and the standard deviation. The results are given in Table 4.

Local collocation knowledge yields the highest accuracy, followed by POS and morphological form. Surrounding words give lower accuracy, perhaps because in our work, only the current sentence forms the surrounding context, which averages about 20 words. Previous work on using the unordered set of surrounding words have used a much larger window, such as the 100-word window of (Yarowsky, 1992), and the 2-sentence context of (Leacock et al., 1993). Verb-object syntactic relation is the weakest knowledge source.

Our experimental finding, that local collocations are the most predictive, agrees with past observation that humans need a narrow window of only a few words to perform WSD (Choueka and Lusignan, 1985).

The processing speed of LEXAS is satisfactory. Running on an SGI Unix workstation, LEXAS can process about 15 examples per second when tested on the "interest" data set.

### 4.2 Evaluation on a Large Data Set

Previous research on WSD tend to be tested only on a dozen number of words, where each word frequently has either two or a few senses. To test the scalability of LEXAS, we have gathered a corpus in which 192,800 word occurrences have been manually tagged with senses from WORDNET 1.5. This data set is almost two orders of magnitude larger in size than the above "interest" data set. Manual tagging was done by university undergraduates majoring in Linguistics, and approximately one man-year of efforts were expended in tagging our data set.

These 192,800 word occurrences consist of 121 nouns and 70 verbs which are the most frequently occurring and most ambiguous words of English. The 121 nouns are:

> action activity age air area art board body book business car case center century change child church city class college community company condition cost country course day death development difference door effect effort end example experience face fact family field figure foot force form girl government ground head history home hour house information interest job land law level life light line man material matter member mind moment money month name nation need number order part party picture place plan point policy pos-ition power pressure problem process program public purpose question reason result right room school section sense service side society stage state step student study surface system table term thing time town type use value voice water way word work world

The 70 verbs are:

add appear ask become believe bring build call carry change come consider continue determine develop draw expect fall give go grow happen help hold indicate involve keep know lead leave lie like live look lose mean meet move need open pay raise read receive remember require return rise run see seem send set show sit speak stand start stop strike take talk tell think turn wait walk want work write

For this set of nouns and verbs, the average number of senses per noun is 7.8, while the average number of senses per verb is 12.0. We draw our sentences containing the occurrences of the 191 words listed above from the combined corpus of the 1 million word Brown corpus and the 2.5 million word Wall Street Journal (WSJ) corpus. For every word in the two lists, up to 1,500 sentences each containing an occurrence of the word are extracted from the combined corpus. In all, there are about 113,000 noun occurrences and about 79,800 verb occurrences. This set of 121 nouns accounts for about 20% of all occurrences of nouns that one expects to encounter in any unrestricted English text. Similarly, about 20% of all verb occurrences in any unrestricted text come from the set of 70 verbs chosen.

We estimate that there are 10-20% errors in our sense-tagged data set. To get an idea of how the sense assignments of our data set compare with those provided by WORDNET linguists in SEMCOR, the sense-tagged subset of Brown corpus prepared by Miller et al. (Miller et al., 1994), we compare a subset of the occurrences that overlap. Out of 5,317 occurrences that overlap, about 57% of the sense assignments in our data set agree with those in SEMCOR. This should not be too surprising, as it is widely believed that sense tagging using the full set of refined senses found in a large dictionary like WORDNET involve making subtle human judgments (Wilks et al., 1990; Bruce and Wiebe, 1994), such that there are many genuine cases where two humans will not agree fully on the best sense assignments.

We evaluated LEXAS on this larger set of noisy, sense-tagged data. We first set aside two subsets for testing. The first test set, named BC50, consists of 7,119 occurrences of the 191 content words that occur in 50 text files of the Brown corpus. The second test set, named WSJ6, consists of 14,139 occurrences of the 191 content words that occur in 6 text files of the WSJ corpus.

| Test set | Sense 1 | Most Frequent | LEXAS |
|---|---|---|---|
| BC50 | 40.5% | 47.1% | 54.0% |
| WSJ6 | 44.8% | 63.7% | 68.6% |

Table 5: Evaluation on a Large Data Set

We compared the classification accuracy of LEXAS against the default strategy of picking the most frequent sense. This default strategy has been advocated as the baseline performance level for comparison with WSD programs (Gale et al., 1992). There are two instantiations of this strategy in our current evaluation. Since WORDNET orders its senses such that sense 1 is the most frequent sense, one possibility is to always pick sense 1 as the best sense assignment. This assignment method does not even need to look at the training sentences. We call this method "Sense 1" in Table 5. Another assignment method is to determine the most frequently occurring sense in the training sentences, and to assign this sense to all test sentences. We call this method "Most Frequent" in Table 5. The accuracy of LEXAS on these two test sets is given in Table 5.

Our results indicate that exemplar-based classification of word senses scales up quite well when tested on a large set of words. The classification accuracy of LEXAS is always better than the default strategy of picking the most frequent sense. We believe that our result is significant, especially when the training data is noisy, and the words are highly ambiguous with a large number of refined sense distinctions per word.

The accuracy on Brown corpus test files is lower than that achieved on the Wall Street Journal test files, primarily because the Brown corpus consists of texts from a wide variety of genres, including newspaper reports, newspaper editorial, biblical passages, science and mathematics articles, general fiction, romance story, humor, etc. It is harder to disambiguate words coming from such a wide variety of texts.

## 5 Related Work

There is now a large body of past work on WSD. Early work on WSD, such as (Kelly and Stone, 1975; Hirst, 1987) used hand-coding of knowledge to perform WSD. The knowledge acquisition process is laborious. In contrast, LEXAS learns from tagged sen-tences, without human engineering of complex rules.

The recent emphasis on corpus based NLP has resulted in much work on WSD of unconstrained real-world texts. One line of research focuses on the use of the knowledge contained in a machine-readable dictionary to perform WSD, such as (Wilks et al., 1990; Luk, 1995). In contrast, LEXAS uses supervised learning from tagged sentences, which is also the approach taken by most recent work on WSD, including (Bruce and Wiebe, 1994; Miller et al., 1994; Leacock et al., 1993; Yarowsky, 1994; Yarowsky, 1993; Yarowsky, 1992).

The work of (Miller et al., 1994; Leacock et al., 1993; Yarowsky, 1992) used only the unordered set of surrounding words to perform WSD, and they used statistical classifiers, neural networks, or IR-based techniques. The work of (Bruce and Wiebe, 1994) used parts of speech (POS) and morphological form, in addition to surrounding words. However, the POS used are abbreviated POS, and only in a window of $\pm 2$ words. No local collocation knowledge is used. A probabilistic classifier is used in (Bruce and Wiebe, 1994).

That local collocation knowledge provides important clues to WSD is pointed out in (Yarowsky, 1993), although it was demonstrated only on performing binary (or very coarse) sense disambiguation. The work of (Yarowsky, 1994) is perhaps the most similar to our present work. However, his work used decision list to perform classification, in which only the single best disambiguating evidence that matched a target context is used. In contrast, we used exemplar-based learning, where the contributions of all features are summed up and taken into account in coming up with a classification. We also include verb-object syntactic relation as a feature, which is not used in (Yarowsky, 1994). Although the work of (Yarowsky, 1994) can be applied to WSD, the results reported in (Yarowsky, 1994) only dealt with accent restoration, which is a much simpler problem. It is unclear how Yarowsky's method will fare on WSD of a common test data set like the one we used, nor has his method been tested on a large data set with highly ambiguous words tagged with the refined senses of WORDNET.

The work of (Miller et al., 1994) is the only prior work we know of which attempted to evaluate WSD on a large data set and using the refined sense distinction of WORDNET. However, their results show no improvement (in fact a slight degradation in performance) when using surrounding words to perform WSD as compared to the most frequent heuristic. They attributed this to insufficient training data in SEMCOR. In contrast, we adopt a different strategy of collecting the training data set. Instead of tagging every word in a running text, as is done in SEMCOR, we only concentrate on the set of 191 most frequently occurring and most ambiguous words, and collected large enough training data for these words only. This strategy yields better results, as indicated by a better performance of LEXAS compared with the most frequent heuristic on this set of words.

Most recently, Yarowsky used an unsupervised learning procedure to perform WSD (Yarowsky, 1995), although this is only tested on disambiguating words into binary, coarse sense distinction. The effectiveness of unsupervised learning on disambiguating words into the refined sense distinction of WORDNET needs to be further investigated. The work of (McRoy, 1992) pointed out that a diverse set of knowledge sources are important to achieve WSD, but no quantitative evaluation was given on the relative importance of each knowledge source. No previous work has reported any such evaluation either. The work of (Cardie, 1993) used a case-based approach that simultaneously learns part of speech, word sense, and concept activation knowledge, although the method is only tested on domain-specific texts with domain-specific word senses.

## 6 Conclusion

In this paper, we have presented a new approach for WSD using an exemplar based learning algorithm. This approach integrates a diverse set of knowledge sources to disambiguate word sense. When tested on a common data set, our WSD program gives higher classification accuracy than previous work on WSD. When tested on a large, separately collected data set, our program performs better than the default strategy of picking the most frequent sense. To our knowledge, this is the first time that a WSD program has been tested on such a large scale, and yielding results better than the most frequent heuristic on highly ambiguous words with the refined senses of WORDNET.

## 7 Acknowledgements

We would like to thank: Dr Paul Wu for sharing the Brown Corpus and Wall Street Journal Corpus; Dr Christopher Ting for downloading and installing WORDNET and SEMCOR, and for reformatting the corpora; the 12 undergraduates from the Linguistics Program of the National University of Singapore for preparing the sense-tagged corpus; and Prof K. P. Mohanan for his support of the sense-tagging project.